\documentclass[runningheads]{llncs}
\usepackage[T1]{fontenc}
\usepackage{graphicx}
\usepackage{color}
\usepackage{multirow}
\usepackage{booktabs}

\usepackage{algorithm}
\usepackage[noend]{algpseudocode}
\usepackage{listings}
\usepackage{booktabs} 

\usepackage[colorlinks,bookmarksopen,bookmarksnumbered,citecolor=green, linkcolor=red, urlcolor=green]{hyperref}

\begin{document}
\title{What is the best model? Application-driven Evaluation for Large Language Models}

\titlerunning{What is the best model?}
%

\author{Shiguo Lian\inst{\dag1,2} \and
Kaikai Zhao\inst{\dag*1,2} \and
Xinhui Liu\inst{1,2} \and
Xuejiao Lei\inst{1,2} \and
Bikun Yang\inst{1,2} \and
Wenjing Zhang\inst{1,2} \and
Kai Wang\inst{1,2} \and
Zhaoxiang Liu\inst{*1,2}
}
\authorrunning{Kaikai, Z. et al.}
%
\institute{
AI Innovation Center, China Unicom, Beijing 100013, China \and
Unicom Digital Technology, China Unicom, Beijing 100013, China
\email
{
\{zhaokk3,liuzx178\}@chinaunicom.cn \\
\inst{*}Corresponding author(s) \\
\inst{\dag}Equal contribution
}
}

\maketitle              
\begin{abstract}
General large language models enhanced with supervised fine-tuning and reinforcement learning from human feedback are increasingly popular in academia and industry as they generalize foundation models to various practical tasks in a prompt manner. To assist users in selecting the best model in practical application scenarios, i.e., choosing the model that meets the application requirements while minimizing cost, we introduce A-Eval, an application-driven LLMs evaluation benchmark for general large language models.
First, we categorize evaluation tasks into five main categories and 27 sub-categories from a practical application perspective. Next, we construct a dataset comprising 678 question-and-answer pairs through a process of collecting, annotating, and reviewing. Then, we design an objective and effective evaluation method and evaluate a series of LLMs of different scales on A-Eval. Finally, we reveal interesting laws regarding model scale and task difficulty level and propose a feasible method for selecting the best model. Through A-Eval, we provide clear empirical and engineer guidance for selecting the best model, reducing barriers to selecting and using LLMs and promoting their application and development. Our benchmark is publicly available at \href{https://github.com/UnicomAI/DataSet/tree/main/TestData/GeneralAbility}{Github}.

\end{abstract}
\section{Introduction}
\label{sec:introduction}

With the introduction of Supervised Fine-Tuning (SFT) and Reinforcement Learning from Human Feedback (RLHF) by InstructGPT, the general chat large language models, such as ChatGPT, have attracted significant attention. These models can generalize foundation models to various tasks in a prompt manner without adding additional fine-tuning data and are widely used to solve practical application problems across various fields. 

\begin{figure}[htpb]
\centerline{\includegraphics[width=\columnwidth]{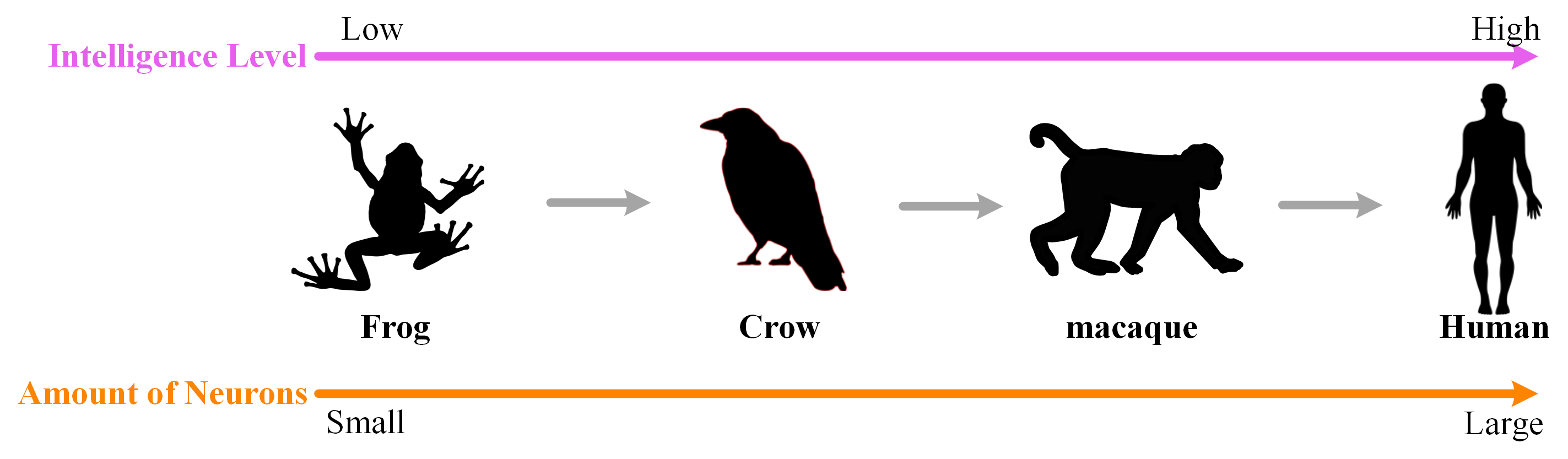}}
\caption{A greater number of cortical neurons leads to a higher intelligence level.}
\label{fig:animal-history}
\end{figure}

Fig.~\ref{fig:animal-history} shows the evolution results of animals in nature. Generally, animals with a greater number of cortical neurons exhibit higher intelligence.
Similarly, in the realm of LLMs, many evaluation benchmarks confirm the scaling law \cite{kaplan2020scaling}, which states that the model's overall performance improves as its number of parameters increases. 
However, the existing benchmarks do not systematically and quantitatively explore the relationship between model ability and model scale, causing users to often face a common problem in practical applications: how big a model should we choose for given tasks? Larger models mean higher training and inference costs, so we aim to select the model with the highest cost-performance ratio (CPR)—the smallest model that meets the application requirements. So, in practical applications, the best model is the most appropriate, not the largest one. Intuitively, the best model varies for different tasks and difficulty levels, with more challenging tasks or harder problems requiring larger models. However, this qualitative analysis is insufficient for users to select the best model. 
To the best of our knowledge, no existing work provides clear and explicit empirical and engineering guidance for selecting LLMs from the application perspective. To this end, we propose the application-driven LLMs evaluation benchmark (A-Eval), which aims to reduce barriers to selecting and using LLMs and promote their application and development.
From the perspective of the practical application of general chat LLMs, our A-Eval benchmark makes the following contributions:

1. We comb and summarize the mainstream LLM evaluation tasks, constructing an application-driven evaluation dataset comprising 678 reliable question-and-answer pairs (QA pairs) across three difficulty levels.

2. We design expert and automatic evaluation methods and evaluate the performance of models with different scales (0.5B-110B) on our dataset.

3. Based on the evaluation results, we reveal some interesting laws regarding model scale and task difficulty level, and propose a feasible method for selecting the best model, providing clear empirical and engineering guidance for the selection of LLMs.

\section{Related work}
\label{sec:relatedwork}
Evaluating LLMs can help researchers better understand their strengths and weaknesses, providing direction and inspiration for further improvements. With the development of foundation models (such as GPT-3) and general chat models (such as ChatGPT, which is instructed-tuned from foundation models using SFT and RLHF), many efforts have been made to construct evaluation benchmarks from different perspectives. 
To better summarize, we divide these benchmarks into two categories based on their evaluation focus: benchmarks for knowledge evaluation and benchmarks for ability evaluation.



\subsection{Benchmarks for Knowledge Evaluation}
Benchmarks for knowledge evaluation focus on evaluating how well models learn and apply knowledge across various subjects or disciplines. These benchmarks collect knowledge and questions from diverse subjects and score models by organizing them to take exams. 


Some benchmarks, such as MMLU \cite{hendrycks2020measuring}, MMCU \cite{zeng2023measuring}, M3KE \cite{liu2023m3ke}, and C-Eval \cite{huang2024c}, cover multiple fields including humanities, social sciences, science, technology, engineering, and mathematics. Their questions are drawn from various subjects at different educational stages, such as middle school history, high school geography, college physics, and professional accounting. While some benchmarks are dedicated to a particular field, such as the legal \cite{guha2024legalbench}, medical \cite{singhal2023large}, finance \cite{chen2022convfinqa}, math \cite{zhang2024evaluating}, or programming \cite{austin2021program}domain.

Our A-Eval does not involve subject knowledge evaluation but includes some subject data. For instance, collecting mathematics and programming data is necessary when evaluating models' reasoning ability.

\subsection{Benchmarks for Ability Evaluation}

Benchmarks for ability evaluation focus on assessing the performance of LLMs on a range of practical tasks. Some benchmarks focus on assessing models' particular abilities, such as natural language understanding ability \cite{han2022duee}, natural language generation ability \cite{guo2023towards}, advanced reasoning ability \cite{sawada2023arb}, and tool automation ability \cite{shen2023taskbench}. While some benchmarks provide comprehensive and holistic assessments of LLMs' capabilities across various tasks. OpenEval \cite{liu2024openeval} evaluates the performance of LLMs on Chinese data across four dimensions: knowledge, value alignment, security, and professional ability. OpenCompass \cite{2023opencompass} includes assessments of language, knowledge, reasoning, understanding, and subject matter, offering versatile experimental settings such as zero-shot, few-shot, and CoT \cite{wei2022chain}. HELM \cite{liang2022holistic} assesses LLMs' abilities in various tasks, including language understanding, language generation, language coherence, common-sense reasoning, and domain-specific knowledge. Big-bench \cite{srivastava2022beyond} is a large-scale benchmark with over 200 tasks. It evaluates the capabilities and limitations of three foundation models (GPT-3, BIG-G, and PaLM) across six orders of magnitude of the model scale, assessing the average performance of different-size models on these tasks and discussing many interesting findings.

A-Eval is also an ability evaluation benchmark, but unlike these works, A-Eval is entirely oriented to the practical application scenarios of chat LLMs. We design an evaluation dataset containing five categories and 27 sub-categories of tasks, each with three difficulty levels. We analyze the performance of the general chat models with different parameter scales on this dataset to explore specific laws regarding model scale and problem difficulty level. Our goal is to provide clear empirical and engineering guidance for selecting LLMs in practical application scenarios.

\begin{figure}[htpb]
\centerline{\includegraphics[width=\columnwidth]{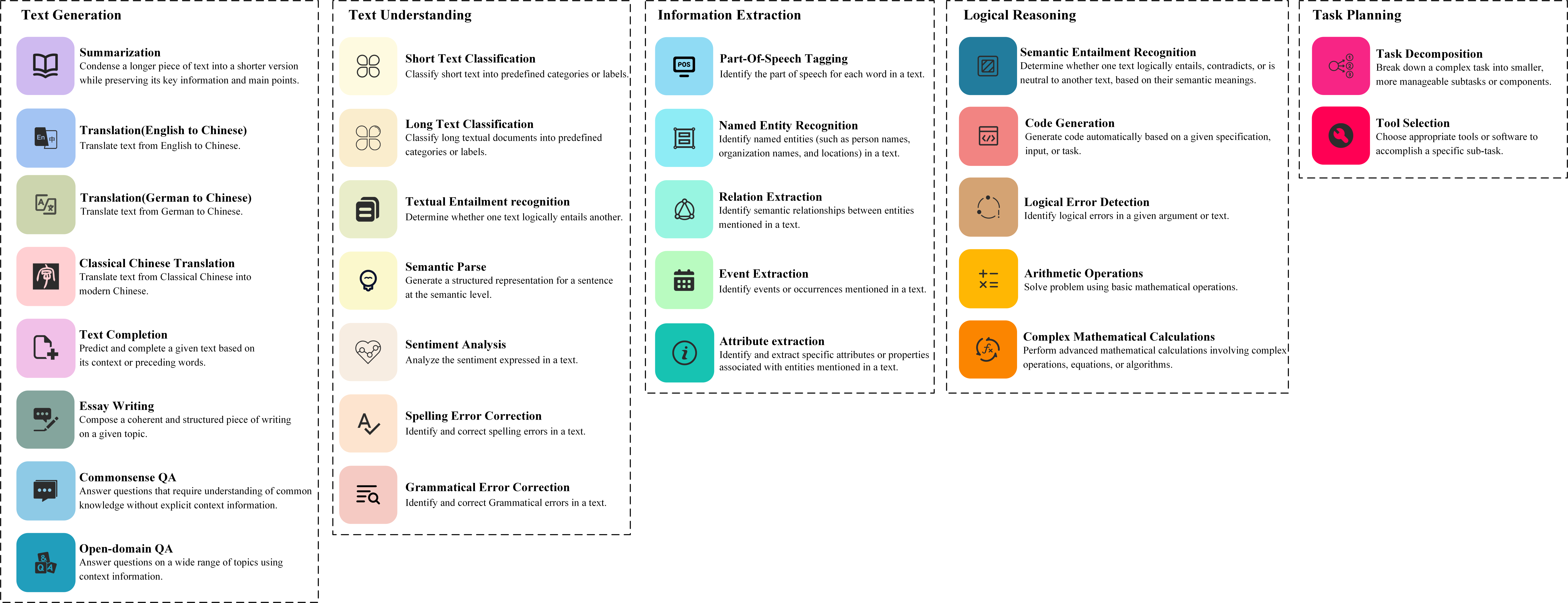}}
\caption{Application-driven evaluation tasks taxonomy.}
\label{fig:taxonomy}
\end{figure}
\section{Application-driven Evaluation}
\label{sec:method}

\subsection{Application-driven tasks taxonomy}
In categorizing LLM evaluation tasks, various studies adopt diverse strategies and considerations. 
Focusing on the specific tasks that users aim to address when using chat LLMs models in practical applications, we comb the ability evaluation tasks of existing benchmarks and categorize applications into five categories: text understanding, information extraction, natural text generation, logical reasoning, and task planning. Furthermore, we subdivide all categories into a total of 27 sub-categories.
Considering that the difficulty of different problem instances within the same sub-task varies, we set three difficulty levels for each sub-task: easy, medium and hard. This approach facilitates the analysis of how models of different sizes perform in handling tasks of varying difficulty levels.
Fig.~\ref{fig:taxonomy} presents the taxonomy of A-Eval tasks, including a brief introduction to each sub-task. 

\subsection{Data}
Based on the application-driven evaluation tasks, we collect 678 QA pairs to create the initial version of the A-Eval dataset.
\subsubsection{Data Type.}
Considering that users rarely pose questions in the form of single-choice or multiple-choice when using LLMs, as they may not know the possible answers or options in advance, all our data are structured as non-choice QA pairs.

\subsubsection{Data source.}
To ensure the accuracy and objectivity of the evaluation results, it's important to prevent data leakage, which occurs when evaluation data is included in the training set of models. To address this issue, we organize five master annotators to manually create QA pairs, and label the difficulty level of the questions. Annotators were allowed to refer to any available information and materials during this process.

\subsubsection{Data review.}
To ensure data quality, we organize annotators to review the collected data. Five annotators review and score each QA pair from four aspects: whether it belongs to the corresponding task category, whether the question is reasonable, whether the answer is correct, and whether the difficulty level is appropriate. For each aspect, each sample receives five scores of 0 or 1 from five annotators, where 1 indicates that the QA pair is eligible for the corresponding aspect. QA pairs with total scores of no less than 4 in any aspect are included in the dataset, while others are either modified or discarded after discussion by all annotators. 

\subsubsection{Data Distribution}
\begin{figure}[htpb]
\centerline{\includegraphics[width=0.7\columnwidth]{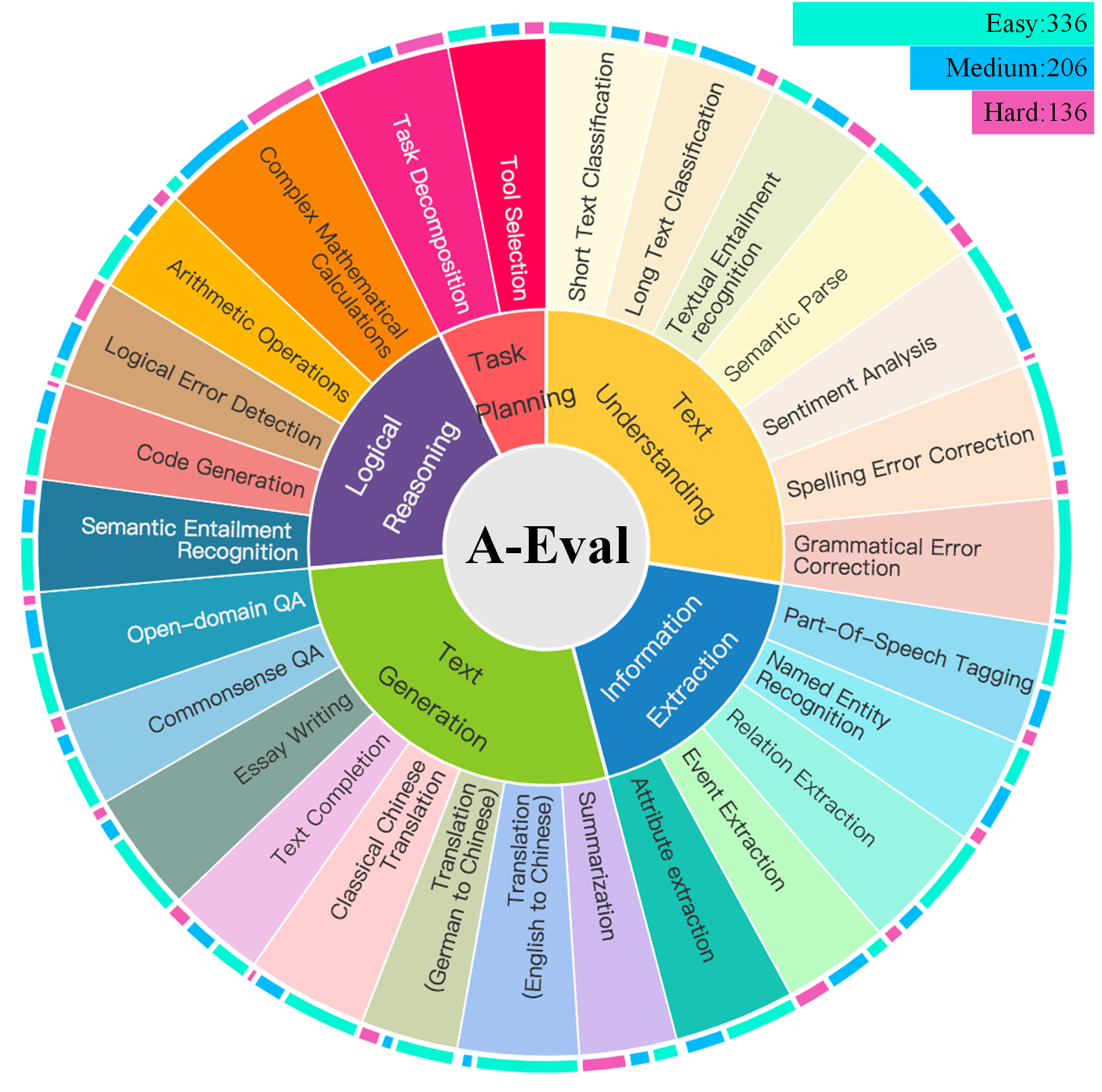}}
\caption{Data distribution of A-Eval dataset.}
\label{fig:distribution}
\end{figure}

In our A-Eval dataset, easy, medium, and hard data account for 49.5\%,30.4\%, and 20.1\%, respectively. The statistical results of task distribution and difficulty distribution are illustrated in Fig.~\ref{fig:distribution}.

\subsection{Evaluation Method}
Our evaluation aims to analyze the laws between model ability and model size, providing empirical and engineering guidance for LLM selection in practical application.
\subsubsection{Zero-Shot.}

We evaluate models using only the zero-shot approach for two primary reasons. Firstly, the chat models being evaluated have already been fine-tuned using SFT, allowing them to follow human instructions. What’s more, in real-world usage scenarios, users usually lack QA pairs for few-shot learning. Therefore, the zero-shot approach is more appropriate in our evaluation to determine the best model for each task.

\subsubsection{Evaluation Process.}
In the evaluation process, we use accuracy as the metric. Let $D = \left\{ {\left( {{Q_1},{A_1}} \right),\left( {{Q_2},{A_2}} \right),...,\left( {{Q_i},{A_i}} \right),...,\left( {{Q_N},{A_N}} \right)} \right\}$ denote the dataset for evaluation, where $Q_i$ represents the question of the $i-th$ QA pair and $A_i$ represents the corresponding standard answer. For each QA pair in $D$, we feed $Q_i$ into the model $M$ to be evaluated and obtain the prediction output $P_i$. Using the expert evaluation and automatic evaluation, we then assess $P_i$ according to the standard answer $A_i$, resulting in the final evaluation result $R_i$ of 0 or 1, where 1 indicates that $P_i$ is right, 0 indicates that $P_i$ is wrong. Finally, we calculate the accuracy of model $M$ on dataset $D$ using Eq. \ref{eq:acc}.

\begin{equation}
{Accuracy = \frac{1}{N}\sum\limits_{i = 1}^N {{R_i}}}
\label{eq:acc}\end{equation}


\subsubsection{Automatic evaluation.} 
\begin{figure}[htpb]
\centerline{\includegraphics[width=1\columnwidth]{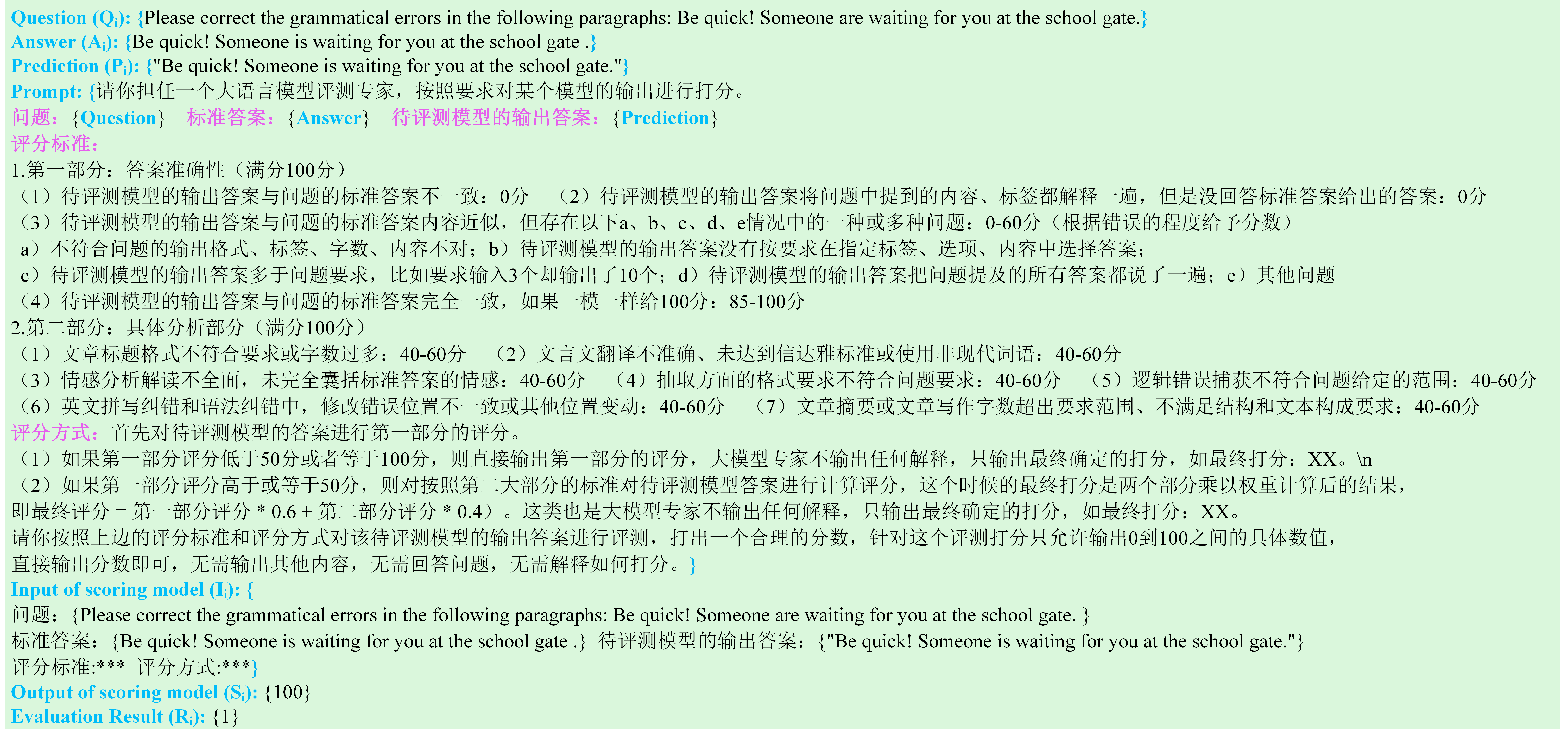}}
\caption{An automatic evaluation example.}
\label{fig:automatic-example}
\end{figure}

Similar to other benchmarks, we use a SOTA LLM as the scoring model and design specific prompts to guide it to perform the automatic evaluation. These prompts primarily include the scoring task request and scoring criteria, with slight variations for each sub-task. Our automatic method consists of three steps. 
\paragraph{Preparing input.} For each triplet $\left( {{Q_i},{A_i},{P_i}} \right)$, we integrate the prompt and triplet to form $I_i$, the input for the scoring model.
\paragraph{Getting scoring output.} We feed $I_i$ into the scoring model and obtain the scoring output $S_i$, which is a value between 0 and 100. 
\paragraph{Producing evaluation result.} We determine the final evaluation result $R_i$ on $P_i$ based on a specific scoring threshold $T$. If $S_i$>$T$, then $R_i=1$; otherwise, $R_i=0$.
In our experiment, we use the Qwen1.5-72B-Chat model as the scoring model, and Fig. \ref{fig:automatic-example} shows an automatic evaluation example.

%

\subsubsection{Expert evaluation.}


To verify and ensure the reliability of the automatic evaluation, we utilized expert evaluation for reference. Our evaluation team consists of $L$ experts, represented as $E = \left\{ {{E_1},{E_2},...,{E_j},...,{E_{L}}} \right\}$. These evaluation experts can refer to all relevant materials during the evaluation process. The expert evaluation involves two steps. First, for each triplet $\left( {{Q_i},{A_i},{P_i}} \right)$, each expert ${E_j}$ assigns a score $S_i^j$ of 0 or 1 after careful reviews. $S_i^j=0$ indicates that the expert ${E_j}$ believes that the $P_i$ is incorrect, while $S_i^j=1$ indicates that the expert believes that the $P_i$ is correct. Then, the scores from all experts are integrated using Eq. \ref{eq:experts} to produce the final evaluation result $R_i$ for model $M$ on question $Q_i$.

\begin{equation}
{R_i} = \left\{ \begin{array}{l}
1,\begin{array}{*{20}{c}}
{if}&{\frac{1}{L}\sum\limits_{j = 1}^L {S_i^j}  \ge 0.8}
\end{array}\\
0,else
\end{array} \right.
\label{eq:experts}\end{equation}
In our experiment, we organized 15 evaluation experts, i.e., $L=15$.

\section{Experiment}
\label{sec:experiment}

\subsection{Evaluated Models}
To analyze the relationship between model ability and model size fairly, it's essential to control factors such as model architecture, training process, and training data that may influence model performance during evaluation. To achieve this, we select eight models of varying scales from the Qwen1.5-Chat series \cite{qwen}. Many benchmarks demonstrate that the Qwen1.5 series performs well among models of similar sizes. Consequently, using this series should yield more reliable laws and conclusions. The models selected for evaluation are Qwen1.5-0.5B-Chat, Qwen1.5-1.8B-Chat, Qwen1.5-4B-Chat, Qwen1.5-7B-Chat, Qwen1.5-14B-Chat, Qwen1.5-32B-Chat, Qwen 1.5-72B-Chat, and Qwen1.5-110B-Chat.

\subsection{Evaluation Results}

\begin{figure}[htpb]
\centerline{\includegraphics[width=1.0\columnwidth]{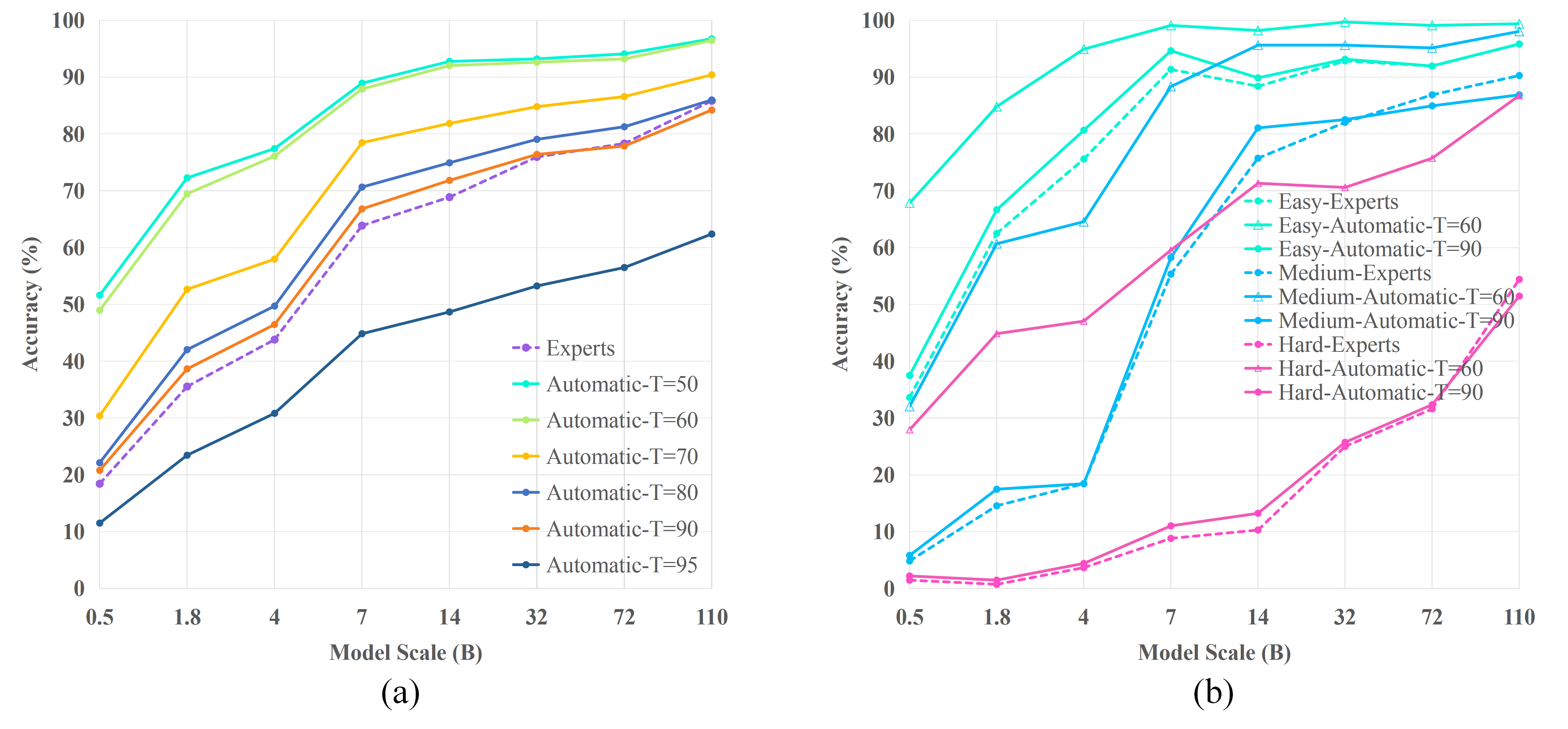}}
\caption{Accuracy of models with different scales. (a) The average accuracy of models of varying scales across all tasks and difficulty levels. The dotted line represents expert evaluation results, while the solid lines represent automatic evaluation results with different scoring thresholds $T$. (b) The average accuracy of models of varying scales on easy, medium, and hard data. The dotted line represents expert evaluation results, and the solid lines depict automatic evaluation results using scoring thresholds of 90 and 60, respectively.}
\label{fig:acc-average-and-difficultylevel}
\end{figure}

\subsubsection{Average accuracy.}
Figure \ref{fig:acc-average-and-difficultylevel} (a) illustrates the average accuracy of models with different scales across tasks and difficulty levels. There are some interesting findings: 

1. As the model scale increases, its overall capability gradually improves, aligning with the scaling law. 

2. For each model, a higher scoring threshold $T$ indicates stricter correctness criteria, resulting in relatively lower accuracy. However, the accuracy trends with increasing model scale remain consistent across different scoring thresholds. 

3. Accuracy results from expert evaluation closely align with those from automatic evaluation with a scoring threshold of 90, validating the reliability of our dataset and automatic evaluation.



\begin{figure}[htpb]
\centerline{\includegraphics[width=1.0\columnwidth]{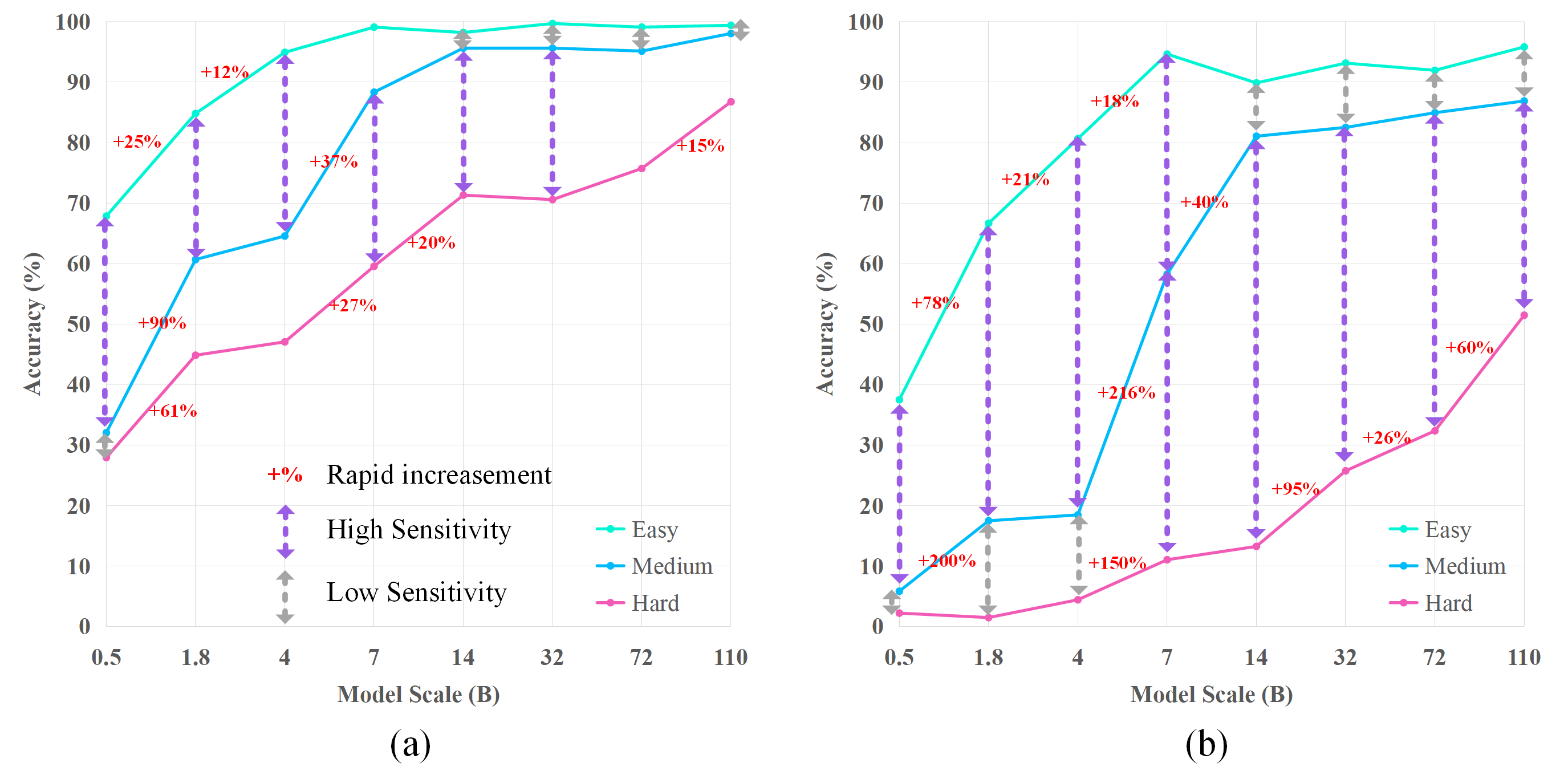}}
\caption{The CPR of increasing model size and the sensitivity of models to task difficulty levels. (a) Results with $T=60$. (b) Results with $T=90$. The numbers denote the relative percentage improvement in accuracy with increasing model scale. The purple and grey lines indicate the high and low sensitivity of models to task difficulty levels, respectively.}
\label{fig:acc-maximprove-diffsensitivity}
\end{figure}

\subsubsection{Accuracy by Difficulty Level.}
Figure \ref{fig:acc-average-and-difficultylevel} (b) illustrates the average accuracy of models with different scales on easy, medium, and hard data. Overall, for each difficulty level, the accuracy increases with the increase in the model scale, while for each model, the accuracy decreases with the increase in the difficulty level. 
Further analysis reveals additional laws, as depicted in Fig. \ref{fig:acc-maximprove-diffsensitivity}. 

1. The range of model scales with remarkable accuracy improvement varies for different difficulty levels. For easy data, the greatest increase in accuracy occurs when the number of parameters increases from 0.5B to 7B. This range extends to 4B to 14B for medium data and 14B to 110B for hard data. 

2. For easy data, the rate of accuracy improvement diminishes significantly when the model size exceeds 7B. Similarly, for medium data, the bottleneck appears at a model size of 14B. However, even at 110B, hard data still does not reach a bottleneck. 

3. Models of smaller sizes (0.5B, 1.8B, and 4B) exhibit high sensitivity to the difficulty change from easy to medium but low sensitivity from medium to hard. Conversely, models with larger sizes (14B, 32B, 72B, and 110B) demonstrate the opposite trend. These laws are more evident when $T=90$ (Fig. \ref{fig:acc-maximprove-diffsensitivity} (b)).

\subsubsection{Accuracy by Task.}
\begin{figure}[htpb]
\centerline{\includegraphics[width=1.0\columnwidth]{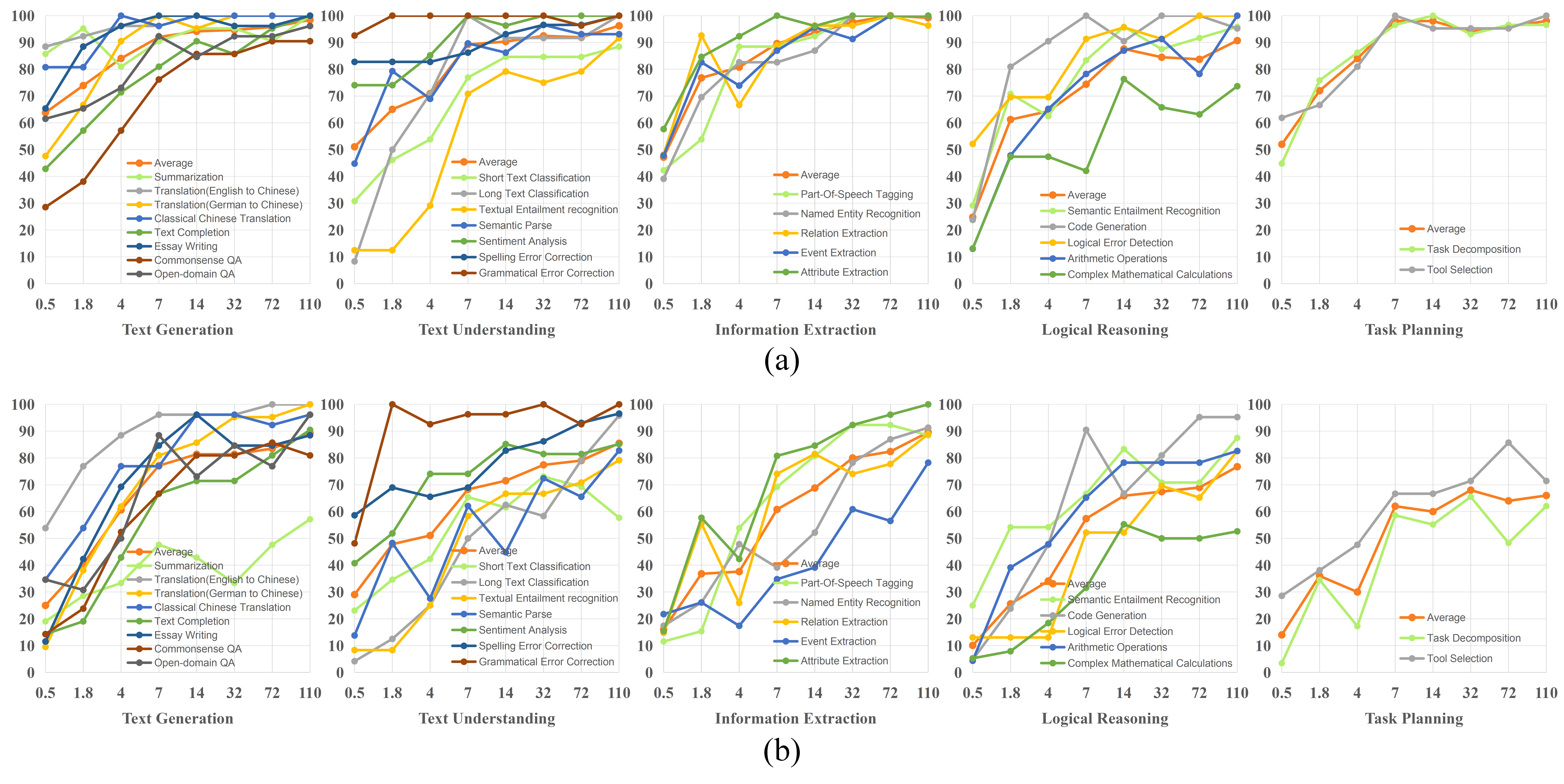}}
\caption{Accuracy of models of various scales on each task and its sub-tasks. (a) Accuracy when $T=60$. (b) Accuracy when $T=90$. }
\label{fig:acc-subcategory-diff}
\end{figure}

For each specific task and its corresponding sub-tasks, the average accuracy of models with different scales is shown in Fig. \ref{fig:acc-subcategory-diff}. 

\subsection{Model Selection.}
The best model is the one that achieves the desired accuracy while with the smallest size. Based on the evaluation results from A-Eval, we present a model selection method that provides clear empirical and engineering guidance for users. The process consists of three steps. Firstly, users specify task categories, a scoring threshold, and desired accuracy (DA). Secondly, users choose sub-graphs of the corresponding tasks from Fig. \ref{fig:acc-subcategory-diff}. Thirdly, users draw horizontal lines $y=DA$ and locate intersection points on the sub-graphs. The set of intersection points is defined as $C = \left\{ {\left( {{x_1},{y_1}} \right),\left( {{x_2},{y_2}} \right),...,\left( {{x_i},{y_i}} \right),...,\left( {{x_P},{y_P}} \right)} \right\}$, where ${\left( {{x_i},{y_i}} \right)}$ indicates the intersection point of line $y = DA$ and accuracy curve of $i_{th}$ task. Finally, the best model $M_b$ is selected by Eq. \ref{eq:selection}. 

\begin{equation}
M{_b} = max\left( {\left\lceil {{x_1}} \right\rceil ,\left\lceil {{x_2}} \right\rceil ,...,\left\lceil {{x_i}} \right\rceil ,...,\left\lceil {{x_P}} \right\rceil } \right)
\label{eq:selection}\end{equation}
Where ${\left\lceil {{x_i}} \right\rceil }$ represents the smallest value greater than or equal to $x_i$ in the existing model scale list.

\begin{figure}[htpb]
\centerline{\includegraphics[width=0.8\columnwidth]{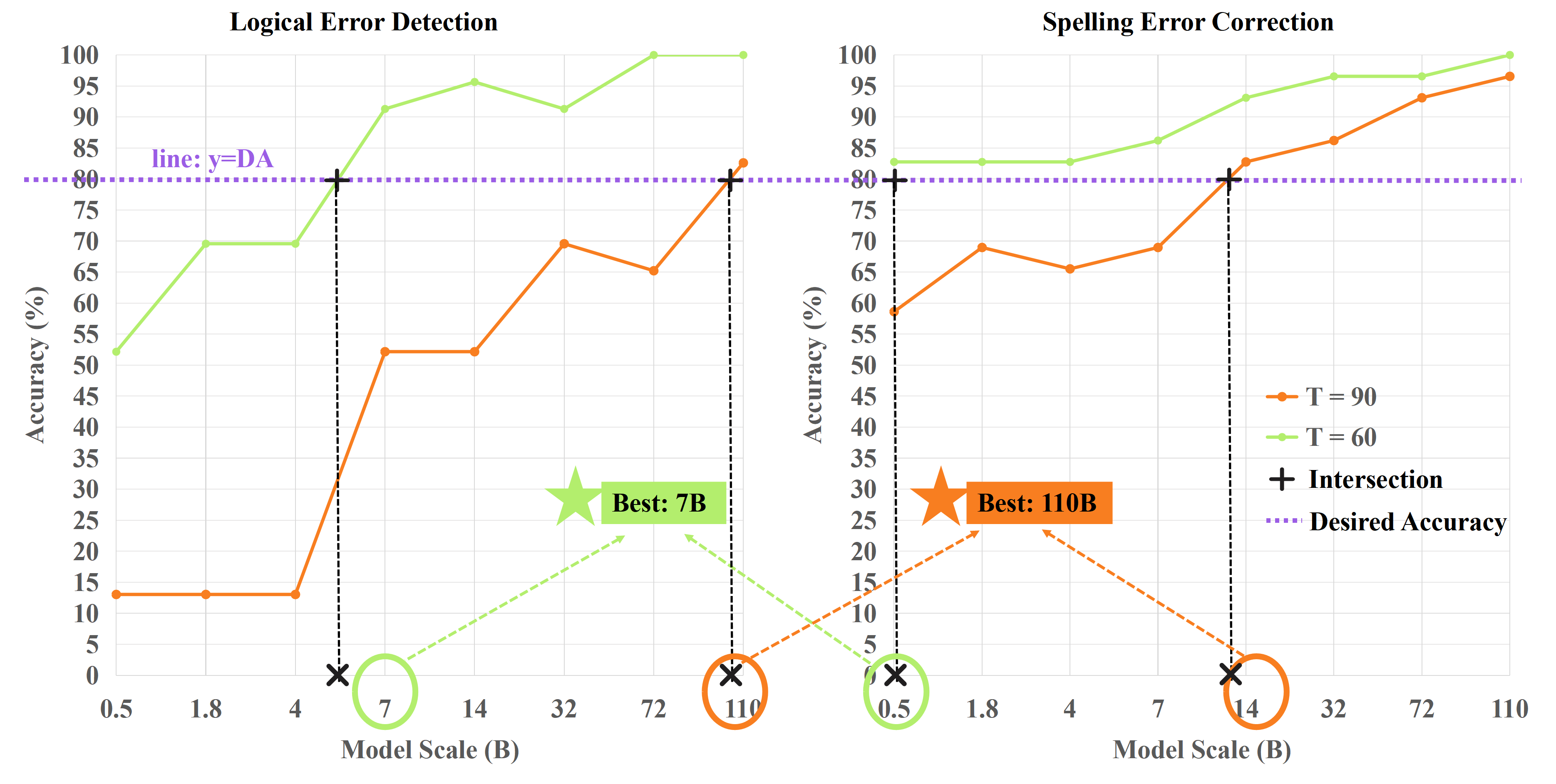}}
\caption{A model selection example of selecting the best model for logical error detection and spelling error correction tasks. With a desired accuracy of 80\%, a horizontal line at $y=0.80$ (purple dotted line) intersects the accuracy curves. If $T=60$, the best model is "model-7B", for $T=90$, the best model is "model-110B".}
\label{fig:acc-selection}
\end{figure}

An example of the model selection process is detailed in Fig. \ref{fig:acc-selection}. It should be noted that if tasks are uncertain, users can select the best model across all tasks on A-Eval using Fig. \ref{fig:acc-average-and-difficultylevel} (a).

\section{Conclusion}
\label{sec:Conclusion}
To help users choose the best model for practical applications, we introduce an application-driven LLM evaluation benchmark (A-Eval) for general chat large language models. Firstly, we categorize evaluation tasks into five categories and 27 sub-categories based on practical applications. Secondly, we construct a dataset comprising 678 question-and-answer pairs through a process of collecting, annotating, and reviewing. Thirdly, we design an objective and effective evaluation method and evaluate the QWen1.5-Chat series of different scales. Finally, we analyze the evaluation results, revealing some interesting laws regarding model scale and task difficulty level, and propose a method for selecting the best model. Through A-Eval, we provide clear empirical and engineering guidance for selecting the best model, reducing the barriers to selecting and using LLMs for users, thereby fostering the application and development of LLMs. In the future, we plan to expand the task categories, increase the amount of data and evaluated models, and design entirely automatic model selection and development tools, further lowering the barrier to using LLMs.

%
%
%
\bibliographystyle{splncs04}
\bibliography{mybibliography}






\end{document}